\documentclass[lettersize,journal]{IEEEtran}
\usepackage{amsmath,amsfonts}
\usepackage{algorithmic}
\usepackage{algorithm}
\usepackage{array}
\usepackage[caption=false,font=normalsize,labelfont=sf,textfont=sf]{subfig}
\usepackage{textcomp}
\usepackage{stfloats}
\usepackage{url}
\usepackage{verbatim}
\usepackage{graphicx}
\usepackage{cite}
\usepackage{float}
\usepackage{hyperref}
\usepackage{xcolor}
\usepackage{booktabs} %需要加载宏包{booktabs}
\usepackage{threeparttable}
\usepackage{multirow} % Required for multirows
\usepackage{bbm}
\begin{document}

\title{DATR: Unsupervised Domain Adaptive Detection Transformer with Dataset-Level Adaptation and Prototypical Alignment}

\author{ 
Jianhong Han$^{\dag}$,~\IEEEmembership{Student Member,~IEEE,}
Liang Chen$^{\dag}$,~\IEEEmembership{Member,~IEEE,}
Yupei Wang$^{*}$,~\IEEEmembership{Member,~IEEE} 

        % <-this % stops a space
% <-this % stops a space
\IEEEcompsocitemizethanks{
\IEEEcompsocthanksitem J. Han, L. Chen and Y. Wang  are with the School of Information and Electronics, Beijing Institute of Technology, Beijing 100081, China, also with the Beijing Institute of Technology Chongqing Innovation Center, Chongqing 401135, China, and also with the National Key Laboratory for Space-Born Intelligent Information Processing, Beijing 100081, China. E-mail: hanjianhong1996@163.com, chenl@bit.edu.cn, wangyupei2019@outlook.com.
}
\thanks{Manuscript submitted to IEEE Transactions on Image Processing.}
}



\maketitle

\begin{abstract}
Object detectors frequently encounter significant performance degradation when confronted with domain gaps between collected data (source domain) and data from real-world applications (target domain). To address this task, numerous unsupervised domain adaptive detectors have been proposed, leveraging carefully designed feature alignment techniques. However, these techniques primarily align instance-level features in a class-agnostic manner, overlooking the differences between extracted features from different categories, which results in only limited improvement. Furthermore, the scope of current alignment modules is often restricted to a limited batch of images, failing to learn the entire dataset-level cues, thereby severely constraining the detector's generalization ability to the target domain. To this end, we introduce a strong DETR-based detector named Domain Adaptive detection TRansformer (DATR) for unsupervised domain adaptation of object detection. Firstly, we propose the Class-wise Prototypes Alignment (CPA) module, which effectively aligns cross-domain features in a class-aware manner by bridging the gap between object detection task and domain adaptation task. Then, the designed Dataset-level Alignment Scheme (DAS) explicitly guides the detector to achieve global representation and enhance inter-class distinguishability of instance-level features across the entire dataset, which spans both domains, by leveraging contrastive learning. Moreover, DATR incorporates a mean-teacher based self-training framework, utilizing pseudo-labels generated by the teacher model to further mitigate domain bias. Extensive experimental results demonstrate superior performance and generalization capabilities of our proposed DATR in multiple domain adaptation scenarios. Code is released at \href{https://github.com/h751410234/DATR}{https://github.com/h751410234/DATR}.
\end{abstract}


\section{Introduction}
\IEEEPARstart{T}{he} domain gap is a common issue encountered during the deployment of deep learning based methods, characterized by distributional discrepancies between collected data (source domain) and data from real-world applications (target domain). Object detectors, which are typically constrained by supervised data-driven architectures, often experience significant performance degradation when confronted with such domain gaps. Due to the high cost and complexity associated with manually annotating data, utilizing unlabeled data from the target domain has increasingly become a practical alternative to address this issue. This situation has spurred the development of unsupervised object detection methods, aimed at mitigating the domain gap through innovative techniques such as adversarial learning\cite{Chen_Li_Sakaridis_Dai_Van_Gool_2018,Saito_Ushiku_Harada_Saenko_2019,Xu_Sun_Yang_Miao_Yang_Research,10070607} and self-training\cite{Liu_Ma_He_Kuo_Chen_Zhang_Wu_Kira_Vajda_2021,Xu_Zhang_Hu_Wang_Wang_Wei_Bai_Liu_2021,10474037}

Recent studies\cite{Wang_Cao_Zhang_He_Zha_Wen_Tao_2021,Huang_Lu_Lin_Xie_Lin,Zhang_Huang_Luo_Zhang_Lu_2021,Yu_Liu_Wei_Zhou_Nakata_Gudovskiy_Okuno_Li_Keutzer_Zhang_2022,Zeng_Ding_Lu} have increasingly shown that the DEtection TRansformer (DETR) exhibits superior performance in addressing domain gaps, outperforming methods based on Convolutional Neural Network (CNN) architectures. Unlike traditional pure convolutional designs, DETR innovatively integrates a CNN backbone with transformer models, i.e., the encoder and the decoder. By utilizing the encoder to further optimize the features extracted from the backbone, DETR significantly improves the representation capability of extracted features\cite{Dosovitskiy_Beyer_Kolesnikov_Weissenborn_Zhai_Unterthiner_Dehghani_Minderer_Heigold_Gelly}. In the decoder, DETR employs multiple object queries to probe local regions and provide instance-level predictions, which effectively simplifies the detection pipeline by eliminating the need for manually designed anchors\cite{Ren_He_Girshick_Sun_2017,Bochkovskiy_Wang_Liao_2020} and Non-Maximum Suppression (NMS)\cite{neubeck2006efficient}. Additionally, the transformer architecture, a crucial component of DETR, has been proven effective in capturing global structural information\cite{Guo_Wang_Qi_Shi_2023,10388356}, substantially boosting the model's generalization capabilities.

Despite the observed gains in accuracy, numerous DETR-based cross-domain detectors\cite{Zhang_Huang_Luo_Zhang_Lu_2021,Geng_Jiang_Shen_Hou_2022} primarily focus on aligning image-level features. These alignment methods employ adversarial learning to obfuscate the origin of domain-specific feature representations, which are developed by integrating features from the backbone network or encoder, thereby enhancing the detectors' ability to extract domain-invariant features. However, these methods often overlook the alignment of instance-level features, a crucial aspect for improving the performance of detectors in cross-domain detection scenarios. Alternatively, some researchers\cite{Gong_Li_Li_Zhang_Liu_Chen_2022,Yu_Liu_Wei_Zhou_Nakata_Gudovskiy_Okuno_Li_Keutzer_Zhang_2022,Wang_Cao_Zhang_He_Zha_Wen_Tao_2021,Tang_Sun_Liu_Yang_2023} attempt to align instance-level features within the decoder, employing methods similar to those used for image-level alignment. These class-agnostic alignment processes, which only distinguish the original domain of features while neglecting the categories they represent, lead to the formation of domain-specific object query representations primarily composed of common foreground features but lose the inherent category information embedded in the object queries, thereby resulting in only limited improvement.

Furthermore, all of the existing approaches are confined to performing feature alignment operations within a limited batch of images, representing only a partial data distribution of the entire dataset. These alignment methods, focused on localized data, impede the model's ability to fully learn and comprehend the complexity and diversity of the entire dataset, thereby constraining the detector's generalization across the target domain.

To address the aforementioned issues, we introduce a strong DETR-based detector named Domain Adaptive Detection Transformer (DATR) for unsupervised domain adaptive object detection. The core design of our detector incorporates two key components: a class-wise prototype alignment module and a dataset-level alignment scheme. The details are explained below:

The motivation of the \textbf{class-wise prototypes alignment} module stems from our ingenious utilization of the prediction mechanism of DETR-based detectors. Within the proposed module, we successfully implement class-aware feature alignment by establishing a bridge between detection tasks and domain adaptation tasks. Specifically, we posit that object queries are capable of predicting specific categories of foreground objects because they aggregate the corresponding semantic information through the cross-attention mechanism in the decoder. Consequently, we directly utilize these prediction results to retroactively ascertain the specific category associated with each object query. Through the proposed efficient batch computation approach, we extract class-wise prototypes in a batch of images with only a single matrix computation. Ultimately, we align these class-wise prototypes by employing adversarial learning to achieve class-aware feature alignment, which significantly improves the detector's performance in cross-domain detection.

The proposed  \textbf{dataset-level alignment scheme} aims to further improve the cross-domain detection performance by aligning features at the dataset level through contrastive learning. The insight behind this design lies in the following two aspects: firstly, the process of building prototypes within a batch of images overlooks dataset-level semantic information, limiting their global representations. Secondly, adversarial learning-based domain adaptation methods, which use a discriminator for binary classification, neglect to optimize the inter-class discriminability of prototypes. Specifically, our scheme stores the class-wise prototypes of both domains in each iteration using a memory module. Then, the model performs cross-domain modeling by computing a strict statistical mean for the stored prototypes. As the stored prototypes grow, the model can develop dataset-level prototypes that reflect dataset-level distribution information rather than the feature representations themselves. Finally, contrastive learning is introduced into the domain adaptation process to bridge the gap between dataset-level prototypes and class-wise prototypes across the two domains. By attracting prototypes of the same class and repelling those of different classes in both domains, the global representation and inter-class distinguishability of instance-level features are effectively improved.

Moreover, DATR integrates the proposed alignment modules with the mean-teacher self-training framework. On one hand, by employing feature alignment methods based on adversarial and contrastive learning, the detector excels in domain-invariant feature extraction and utilization. On the other hand, the construction of a self-training framework generates pseudo-labels for target domain images. Leveraging these pseudo-labels, the domain bias of the detector is further mitigated through supervised learning.

The main contributions of this paper are as follows:

\begin{itemize}
\item{We develop a Class-wise Prototypes Alignment module (CPA) for class-aware feature alignment by bridging the gap between object detection task and domain adaptation tasks, which significantly improves the detector's performance in cross-domain detection.}

\item{The Dataset-level Alignment Scheme (DAS) is proposed for feature alignment across the entire dataset using contrastive learning, which achieves global representation and enhances inter-class distinguishability of instance-level features.}

\item{We show that DETR-based detectors can be effectively combined with a self-training framework for cross-domain detection tasks. This combination can further mitigate the domain bias of the detector by leveraging the generated pseudo-labels.}

\end{itemize}

Extensive results demonstrate the superior performance and generalization capabilities of DATR compared to state-of-the-art methods in multiple adaptation scenarios. The Weather Adaptation (Cityscapes\cite{Cordts_Omran_Ramos_Rehfeld_Enzweiler_Benenson_Franke_Roth_Schiele_2016} → Foggy Cityscapes\cite{Sakaridis_Dai_Van_Gool_2018}) resulted in a notable improvement of over \begin{math} 52.8\% \end{math} in mean Average Precision (mAP). The Synthetic-to-Real Adaptation (Sim10k\cite{Johnson_Roberson_Barto_Mehta_Sridhar_Rosaen_Vasudevan_2016} → Cityscapes) demonstrated remarkable enhancements, with mAP increases exceeding \begin{math} 66.3\% \end{math}. The Scene Adaptation from (Cityscapes → BDD100k\cite{yu2020bdd100k}) achieved a \begin{math} 41.9\% \end{math} in mAP.

\section{Related Work}
\subsection{Unsupervised Domain Adaptive Object Detection}

The objective of Unsupervised Domain Adaptation (UDA) is to mitigate domain gap by utilizing unlabeled data from target domains. In object detection, Chen et al.\cite{Chen_Li_Sakaridis_Dai_Van_Gool_2018} pioneered a UDA approach based on the Faster R-CNN detector, which is foundational for subsequent developments. This approach encompasses dual-feature alignment: image-level and instance-level, utilizing domain discriminators for adversarial training to enable different components of the detector to extract domain-invariant features. Saito et al.\cite{Saito_Ushiku_Harada_Saenko_2019} observed that drastic changes in detection scenarios, such as scene layouts or object counts, can adversely affect the accuracy of feature alignment. Consequently, they introduced a hybrid method combining weak global with strong local alignment to extract domain-invariant features more effectively. Chen et al.\cite{Zhu_Pang_Yang_Shi_Lin_2019} underscored the significance of local object regions in detection and developed an advanced instance-level alignment module by employing K-means clustering to identify candidate regions that are in closer proximity.

With the significant success of DETR, it has attracted substantial attention from researchers exploring its potential for domain adaptation tasks. Wang et al.\cite{Wang_Cao_Zhang_He_Zha_Wen_Tao_2021} proposed SFA, introducing two alignment methods for transformer-based detectors: query-based and token-wise feature alignments. Huang et al.\cite{Huang_Lu_Lin_Xie_Lin} innovatively integrated adversarial feature alignment into detection transformers, introducing an adversarial token mechanism alongside cross-attention layers. Zhang et al.\cite{zhang2023detr} presented the CNN-Transformer Blender (CTBlender), an ingeniously fusion of CNN and Transformer features, to enhance feature alignment in the backbone and encoder of detection models. Differently, we explore a strong DETR-based detector for unsupervised domain adaptive object detection, which enables effective instance-level feature alignment across categories.

\subsection{Contrastive Learning}

Contrastive learning\cite{Hadsell_Chopra_LeCun_2006,He_Fan_Wu_Xie_Girshick_2020,Chen_Kornblith_Norouzi_Hinton_2020,10482848} has emerged as a highly effective approach in the field of self-supervised representation learning. This method enhances feature representation by contrasting pairs of data samples, ensuring that representations of similar samples are attracted, while those of dissimilar samples are distinctly repelled. Recently, some researchers have attempted to apply contrastive learning to unsupervised domain adaptation tasks, achieving significant progress. CLST\cite{Marsden_Bartler_Dobler_Yang_2022} employs contrastive learning to refine domain-invariant feature representations, thereby enabling sophisticated unsupervised cross-domain semantic segmentation. ProCA\cite{Jiang_Li_Yang_Gao_Wang_Tai_Wang_2022} leverages contrastive learning to explicitly model the relationships of pixel-wise features between different categories and domains, achieving strong domain-invariant representations in unsupervised domain adaptive semantic segmentation. Moreover, CMT\cite{Cao_Joshi_Gui_Wang_2023} explores the synergy between the Mean Teacher model\cite{Tarvainen_Valpola_2017} and contrastive learning, effectively achieving cross-domain object detection. In this paper, our proposed dataset-level alignment scheme successfully achieves feature alignment across the entire dataset using contrastive learning. This scheme improves the global representation and inter-class discriminability of instance-level features.

\subsection{Self-training Framework}

Self-training frameworks have become a pivotal strategy in semi-supervised object detection tasks and have gained widespread application. These frameworks utilize unlabeled data to generate pseudo-labels for supervised training, thereby significantly reducing reliance on costly annotated data during the training phase. The Unbiased Teacher methodology\cite{Liu_Ma_He_Kuo_Chen_Zhang_Wu_Kira_Vajda_2021}, which introduces a novel student-teacher mutual learning pipeline for pseudo-label generation, has been extensively adopted in subsequent research. The Soft Teacher approach\cite{Xu_Zhang_Hu_Wang_Wang_Wei_Bai_Liu_2021} innovatively employs pseudo-label scores as weights in loss calculations, thereby enhancing the accuracy of the pseudo-labels. To address the issue of label mismatch, LabelMatch\cite{Chen_Chen_Yang_Xuan_Song_Xie_Pu_Song_Zhuang_2022} skillfully employs label distribution to dynamically determine filtering thresholds for different pseudo-label categories. 

Recently, AT\cite{Li_Dai_Ma_Liu_Chen_Wu_He_Kitani_Vajda_University_et}, have demonstrated the effectiveness of applying self-training frameworks to cross-domain adaptation tasks, effectively combining them with domain adversarial learning to bridge domain gaps. MTTrans\cite{Yu_Liu_Wei_Zhou_Nakata_Gudovskiy_Okuno_Li_Keutzer_Zhang_2022} transplants the mean teacher framework\cite{Tarvainen_Valpola_2017} onto Deformable DETR, leveraging pseudo-labels in object detection training to facilitate knowledge transfer between domains. We integrate DINO\cite{zhang2022dino} into the mean-teacher self-training framework, and this combination can further mitigate the domain bias of the detector by leveraging the generated pseudo-labels for supervised learning.

\section{Method}

\begin{figure*}[htbp]
\centering
\includegraphics[width=18cm]{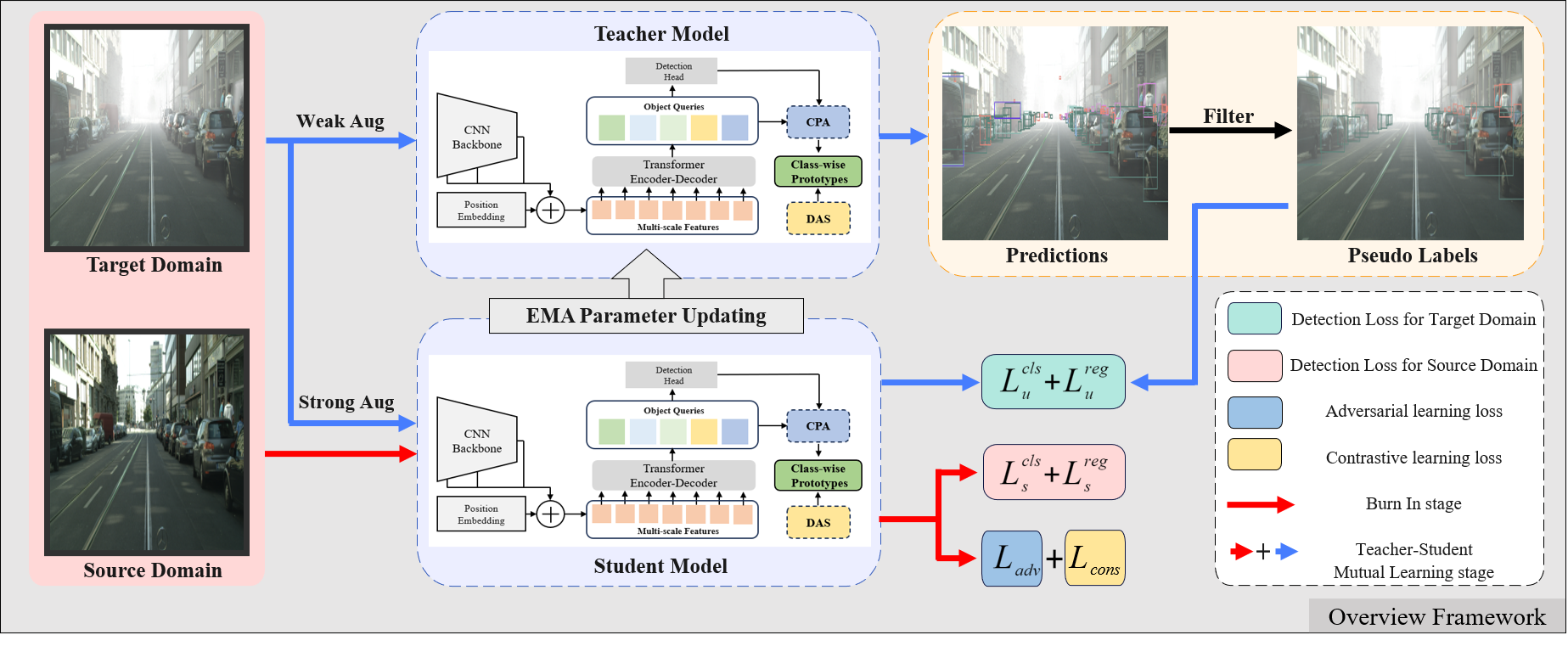}
\caption{
The DATR employs a self-training framework that includes two models: a student model, serving as the core task model, and its temporally ensembled counterpart, known as the teacher model. The student model effectively aligns cross-domain features in a class-aware manner by utilizing the proposed Class-wise Prototypes Alignment (CPA) module. Subsequently, the designed Dataset-level Alignment Scheme (DAS) assists the detector in enhancing cross-domain feature alignment across the entire dataset through the use of contrastive learning. The teacher model, updated by the EMA of the student model, generates pseudo labels for images in the target domain. DATR utilizes these pseudo-labels to further mitigate the domain bias within the detector. We divide the training process into two stages. In the Burn-In stage, we exclusively train the student model, incorporating both supervised and unsupervised learning. In the Teacher-Student Mutual Learning stage, unlabeled data from the target domain are fed into the teacher model to generate pseudo labels for supervised learning.
}
\label{fig_1}
\end{figure*}

\subsection{Overview}

\textbf{DETR revisit.} The DEtection TRansformer (DETR) is a transformer-based, end-to-end object detector that eliminates the need for conventional hand-designed components, such as anchor design and non-maximum suppression (NMS). Specifically, DETR employs a CNN backbone for feature extraction, subsequently feeding into an encoder-decoder transformer structure and a Feed-Forward Network (FFN) for final detection predictions. Building on the foundation of DETR, DINO\cite{zhang2022dino} enhances performance by aggregating multi-scale features and improving object queries initialization for more accurate predictions. Furthermore, it employs denoising training to accelerate convergence speed.

\textbf{Framework overview.} As shown in Fig. \ref{fig_1}, our DATR employs an iterative teacher-student learning framework that consists of two components with identical architectures: a student model and a teacher model. Each model adopts DINO as the base detector and integrates our designed methods, including the Class-wise Prototypes Alignment (CPA) module and the Dataset-level Alignment Scheme (DAS). Following the existing methods \cite{10474037,Liu_Ma_He_Kuo_Chen_Zhang_Wu_Kira_Vajda_2021,Xu_Chen_Guan_Hu_Group}, we divide the training process into two stages: the Burn-In stage and the Teacher-Student Mutual Learning stage.
In the Burn-In stage, we exclusively train the student model, incorporating both supervised and unsupervised learning, the pipeline is illustrated in Fig. \ref{fig_2} (a). Specifically, the network processes pairs of images, containing an equal number of images from both the source and target domains. For each image pair, DATR employs a ResNet backbone\cite{He_Zhang_Ren_Sun_2016} to extract features. Subsequently, these features are enhanced and decoded by the transformer's encoder-decoder architecture, yielding object queries that probe local regions and aggregate instance-level features. Ultimately, the detector computes the predictions by processing these object queries through a 3-layer MLP equipped with a ReLU activation function. For supervised learning, we exclusively utilize predictions from the source domain to calculate the detection loss \begin{math} L_{det} \end{math}, similar to DINO, due to the absence of pseudo-label generation at this stage. For unsupervised learning, we introduce the Class-wise Prototypes Alignment (CPA) module that effectively aligns cross-domain features in a class-aware manner by bridging the gap between detection tasks and domain adaptation tasks. More details of CPA module are available in Subsection \ref{Subsection B}. Moreover, we develop a Dataset-level Alignment Scheme (DAS) that employs contrastive learning to align features across the entire dataset, which enhances the global representation and inter-class discriminability of instance-level features. The DAS will be presented in Subsection \ref{Subsection C}.

In the Teacher-Student Mutual Learning stage, the teacher model is integrated into the training framework to help the detector further mitigate domain bias. Specifically, the approach used in unsupervised learning is identical to that of the Burn-In stage. Differently, in this stage, unlabeled data from the target domain are introduced into the teacher model to generate pseudo-labels for supervised learning. During training, the student model is updated by minimizing the loss through gradient descent. Following\cite{Mi_Lin_Zhou_Shen_Luo_Sun_Cao_Fu_Xu_Ji}, the parameters of the teacher network are updated from the student network via Exponential Moving Average (EMA)\cite{Marsella_Gratch_2009}, as follows:
\begin{equation}
\begin{split}
\theta_{t}^{i} \leftarrow \alpha \theta_{t}^{i-1}+(1-\alpha) \theta_{s}^{i},
\end{split}
\end{equation}
where \begin{math} \theta_{t} \end{math} and \begin{math} \theta_{s} \end{math} are the parameters of the teacher and student networks, respectively, and \begin{math} {i} \end{math} denotes the training step. \begin{math} \alpha \end{math} is the hyper-parameter to determine the speed of parameter transmission, which is normally close to 1.

\subsection{Class-wise Prototypes Alignment Module}
\label{Subsection B}

Here, DATR effectively aligns cross-domain features in a class-aware manner by utilizing the Class-wise Prototypes Alignment (CPA) module, as detailed below:

%待添加背景故事+概述
\textbf{Class-wise prototypes extraction.} We innovatively establish a connection between domain adaptation tasks and detection tasks by leveraging the detection results to extract prototypes on a class-wise basis. Specifically, upon receiving an output object query \begin{math} Z_n\in\mathbb{R}^{(N\times d)} \end{math}  from the decoder, we follow the detection pipeline to determine its predicted category \begin{math} C_n \end{math}. Subsequently, we merge the features of object queries within the same category and compute the centroids of these aggregated features to obtain the class-wise prototypes \begin{math} P_c\in\mathbb{R}^{(C\times d)} \end{math}, as illustrated in the following formula:

\begin{equation}
P_{c}=\frac{\sum_{n=1}^{N} Z_{n} \mathbbm{1}\left[C_{n}=c\right]}{\sum_{n=1}^{N} \mathbbm{1}\left[C_{n}=c\right]},
\end{equation}
where \begin{math} Z_n \end{math} represents the learned embeddings that decode object representations from the output of the encoder, with a dimension of \begin{math} d \end{math}. The variable \begin{math} c \end{math} denotes the index corresponding to one of the total categories. The function \begin{math} \mathbbm{l}[C_n=c] \end{math} serves as an indicator, equalling 1 when \begin{math} C_n=c \end{math} and 0 otherwise. Class-wise prototypes \begin{math} P_c \end{math} are considered the approximate representational centroids of the various categories.

\textbf{Adversarial learning to align class-wise prototypes.} We obtain class-wise prototypes from two distinct domains through feature aggregation of object queries. Adversarial learning is employed to align the feature representations of these prototypes across both domains. Specifically, class-wise prototypes are fed into a simple CNN-based discriminator \begin{math} D \end{math} to determine a probability that indicates their origin (either source or target domain). Prototypes originating from the source domain are labeled as \begin{math} d=0 \end{math}, while those from the target domain are labeled as \begin{math} d=1 \end{math}. This enables us to optimize objectives using binary cross-entropy loss as :
\begin{equation}
L_{a d v} = -\sum_{N}\left[d \log p_{(N)}+(1-d) \log \left(1-p_{(N)}\right)\right],
\end{equation}
where \begin{math} p_{(N)} \end{math} represents the output of the discriminator. We implement end-to-end adversarial learning by integrating a Gradient Reversal Layer (GRL)\cite{Ganin_Lempitsky_2015}. During training, the class-wise prototypes aim to deceive the discriminator, which is tasked with discerning the domain of origin of these prototypes. Consequently, the object queries adapt to utilize domain-invariant features extracted by the encoder to obtain detection results. Thus, the adversarial optimization objective function is defined as follows:
\begin{equation}
L_{a d v}=\max _{P} \min _{D} L_{d i s},
\end{equation}
where \begin{math} P \end{math} represents the class-wise prototypes, and \begin{math} D \end{math} denotes the simple CNN-based discriminator.

\begin{figure*}[t]
\centering
\includegraphics[width=19cm]{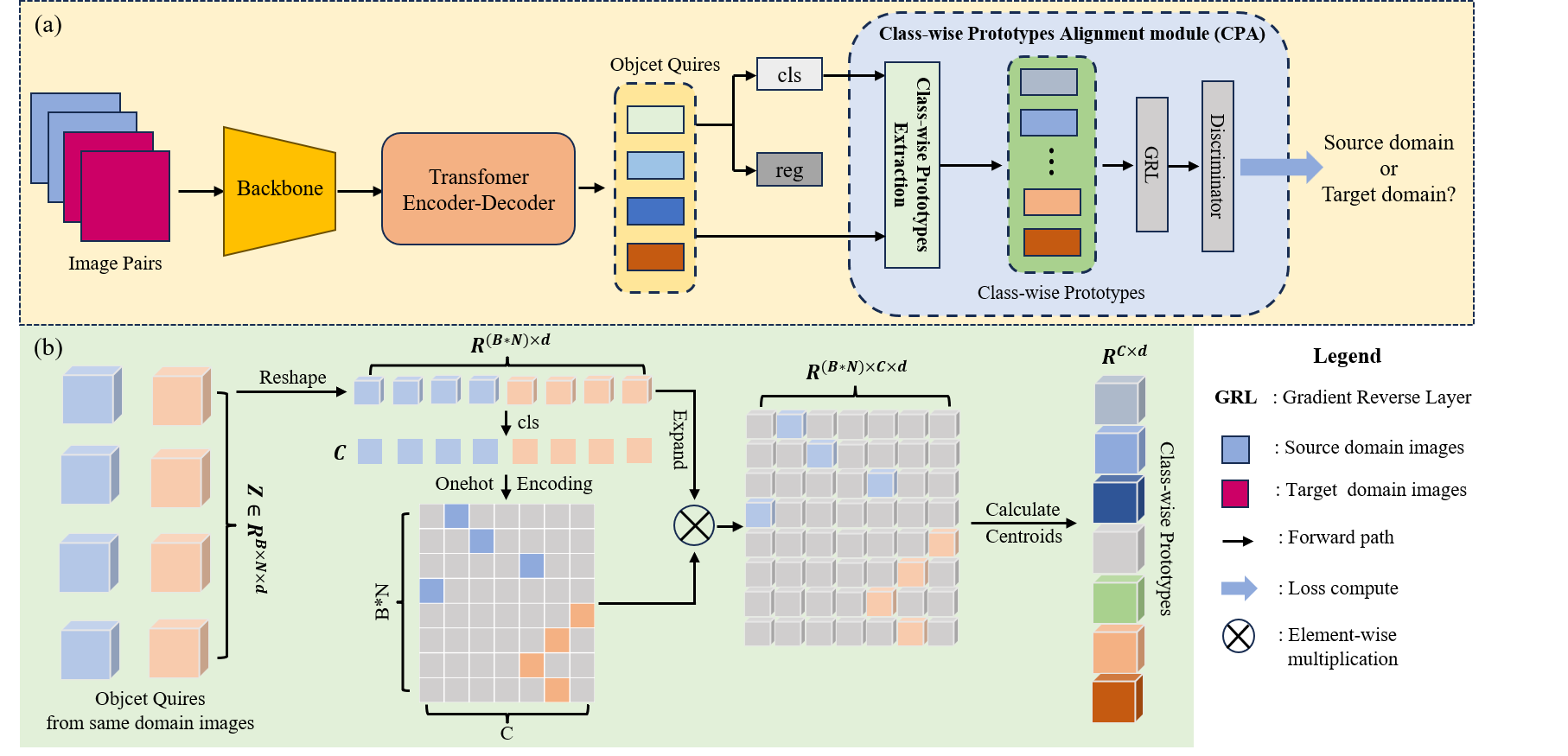}
\caption{Details of (a) the proposed detection pipeline, which incorporates the Class-wise Prototypes Alignment (CPA) module for achieving cross-domain feature alignment, and (b) the efficient batch computation method for extracting class-wise prototypes through the use of class masks.}
\label{fig_2}
\end{figure*}

\textbf{Efficient batch computation with class mask.} An intuitive method for extracting prototypes on a class-wise basis involves conducting iterative calculations across each category within a batch of images, a process that significantly impedes the efficiency of network training. To address this challenge, we propose a more efficient batch computation approach by utilizing a strategically designed class mask, as illustrated in Fig. \ref{fig_2} (b). In a given training batch, object queries are defined as \begin{math} \mathbf{Z}\in\mathbb{R}^{(B\times N\times d)} =\left \{ [Z_1^1,Z_2^1,\ldots,Z_N^1],[Z_1^2,Z_2^2,\ldots,Z_N^2],\ldots,[Z_1^i,Z_2^i,\ldots,Z_N^i] \right \} \end{math}, where each \begin{math} Z_j^i \end{math} represents an object query, and \begin{math}N \end{math} signifies the total number of object queries for the \begin{math} i-th \end{math} image. The corresponding classification outcomes, denoted as \begin{math} \mathbf{C}\in\mathbb{R}^{(B\times N\times 1)} =\left \{ [C_1^1,C_2^1,\ldots,C_N^1],[C_1^2,C_2^2,\ldots,C_N^2],\ldots,[C_1^i,C_2^i,\ldots,C_N^i] \right \} \end{math}, are ascertained through the detection head, with each \begin{math} C_j^i \end{math} representing the predicted category for the corresponding \begin{math} Z_j^i \end{math}. We commence by reshaping object queries \begin{math} \mathbf{Z} \end{math} and their corresponding classification results \begin{math} \mathbf{C} \end{math}, streamlining the process to derive the outcomes via a single matrix computation. Subsequently, we convert the classification results into a class mask using one-hot encoding, which effectively indexes and tracks the relevant object queries. In the final step, we exploit the broadcasting mechanism inherent in matrix multiplication, enabling efficient computation of centroids and thereby facilitating the derivation of class-wise prototypes. It is important to highlight that without a class mask, adversarial learning often leads to a reduction in performance rather than an improvement. We posit that the primary reason for this observation is that class masks serve not only to accelerate computation but also to prevent the computation of adversarial losses for non-existent categories within the image.

\textbf{Variants.} We endeavor to further explore the potential of our proposed module, primarily by selecting more representative object queries for the extraction of class-wise prototypes. Diverging from the aforementioned unfiltered approach, we introduce two variants based on different selection criteria. Specifically, the first variant involves selecting reliable object queries based on the confidence value of prediction results. Intuitively, higher confidence in predictions suggests that the aggregated features by object queries are more accurate in representing objects, thus yielding more representative class-wise prototypes. The second variant employs the Hungarian matching algorithm to find object queries that uniquely match annotations, utilizing these queries to derive class-wise prototypes. Compared to the first variant, this method filters out a greater number of object queries, yielding even more representative prototypes. It is important to note that in the target domain, which lacks real annotated labels, we consider employing pseudo-labels as substitutes, equivalent to setting a higher confidence threshold in this domain. Counterintuitively, the variants derived from filtered object queries did not improve cross-domain performance, as detailed in experimental section \ref{Ablation studies}.

\subsection{Dataset-level Alignment Scheme}
\label{Subsection C}

\begin{figure*}[t]
\centering
\includegraphics[width=18cm]{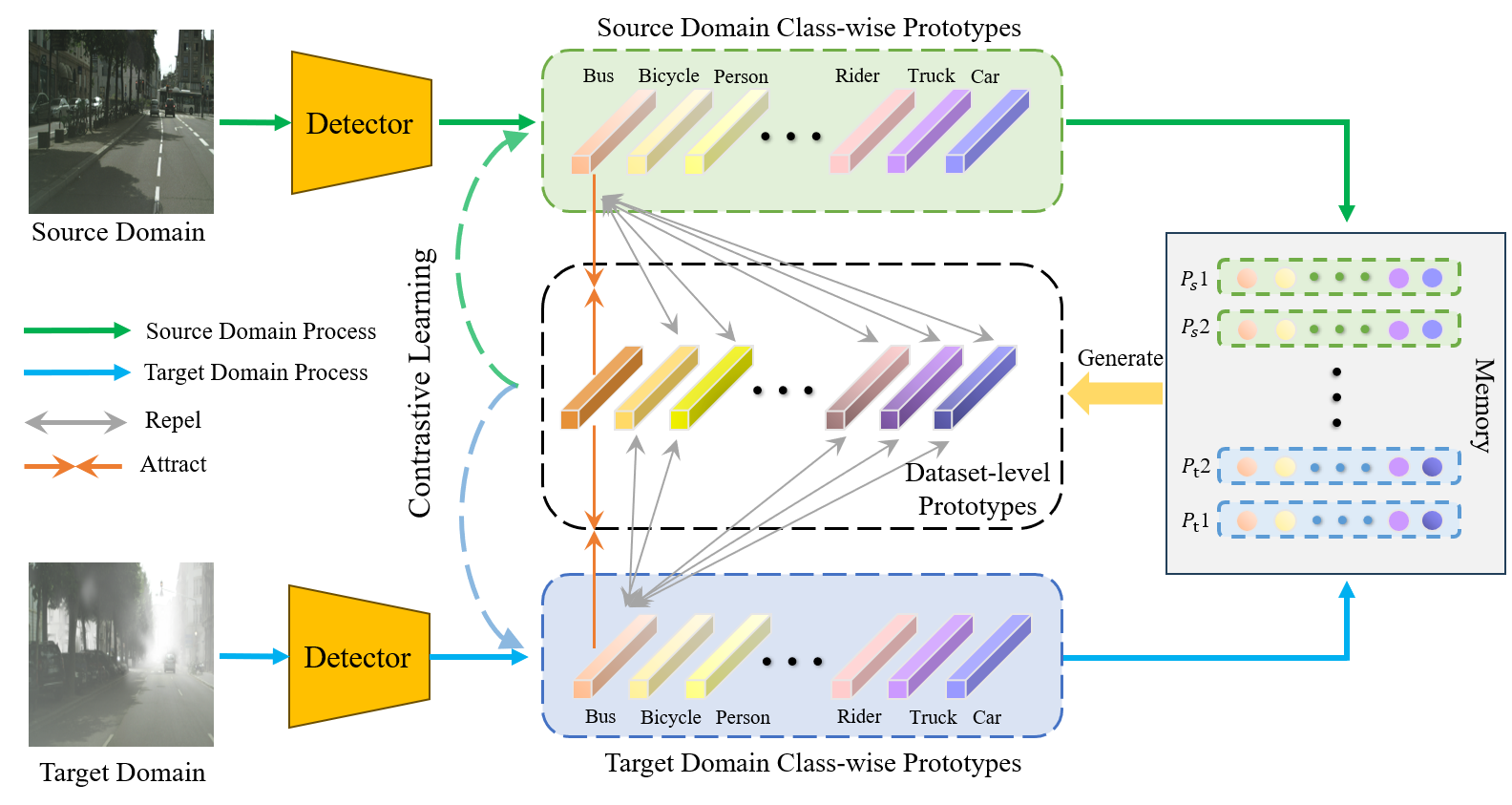}
\caption{
Our proposed Dataset-level Alignment Scheme (DAS). Dataset-level prototypes can be generated using a memory module. Contrastive learning is applied across two domains to enforce refined feature adaptation.
}
\label{fig_3}
\end{figure*}

While the alignment of class-wise prototypes has facilitated the use of domain-invariant features in object queries for object detection, there remains untapped potential for enhancing these extracted prototypes. Firstly, the domain adaptation based on adversarial learning, which employs a discriminator for input origin determination (binary classification), tends to overlook the optimization of inter-class discriminability of the prototypes. Secondly, the aforementioned method focuses only on aligning prototypes within a batch of images, thereby neglecting valuable contextual information at the dataset level. This oversight can limit the potential for the global representation of class-wise prototypes. To address the aforementioned challenges, we have developed a dataset-level alignment scheme. The proposed scheme constructs dataset-level prototypes across various domains by leveraging the intuitive principle of visual consistency within the same categories. We perform contrastive learning between dataset-level prototypes and class-wise prototypes to further enhance the global representation and inter-class discriminability of instance-level features.

\textbf{Cross-domain dataset-level  prototypes aggregation.} As illustrated in Fig. \ref{fig_3}, we use a memory module to store class-wise prototypes extracted in each iteration and model them as dataset-level representations. In this work, when generating these dataset-level representations, we compute the strict statistical mean of the stored prototypes as follows: 

\begin{equation}
{\widetilde{P}}_c=\frac{P_cn_{c+}{\widetilde{P}}_c{\widetilde{n}}_c}{n_c+{\widetilde{n}}_c},
\end{equation}
where \begin{math} P_c \end{math} represents the class-wise prototypes as estimated online, and \begin{math} n_c \end{math} indicates the total count of object queries belonging to category \begin{math} c \end{math} in a newly added mini-batch during training. \begin{math} {\widetilde{P}}_c \end{math} corresponds to the dataset-level representations generated by the memory module, which is initially set to \begin{math} 0 \end{math}. \begin{math} {\widetilde{n}}_c \end{math} denotes the cumulative number of object queries associated with category \begin{math} c \end{math} up to the last update. 

During the training process, \begin{math} P_c \end{math} extracted from both the source and target domains, is utilized to update the same dataset-level representations \begin{math} {\widetilde{P}}_c \end{math}. This scheme of mixing prototypes could be regarded as a bridge connecting the two domains by leveraging the intuitive principle of visual consistency. Ultimately, the resulting \begin{math} {\widetilde{P}}_c \end{math} is naturally employed in cross-domain tasks.

\textbf{Contrastive learning for domain adaptation.} We perform contrastive learning between dataset-level representations and class-wise prototypes. By leveraging the optimization mechanism of contrastive learning, where positive pairs are attracted to each other and negative pairs are repelled, we further enhance the global representation and inter-class discriminability of instance-level features. Specifically, we define \begin{math} P_c^S\in\mathbb{R}^{C\times d} \end{math} and \begin{math} P_c^T\in\mathbb{R}^{C\times d} \end{math} as the class-wise prototypes from the source and target domains, respectively. \begin{math}{\widetilde{P}}_c\in\mathbb{R}^{C\times d} \end{math} represents the dataset-level representations. We engage in contrastive learning between \begin{math} {\widetilde{P}}_c \end{math} and \begin{math} P_c^S \end{math}, as well as between \begin{math} {\widetilde{P}}_c \end{math} and \begin{math} P_c^T \end{math}, treating features of the same category as positive samples and others as negative samples. The contrastive loss is formulated as:
\begin{equation}
\begin{aligned}
L_{\text {contrast }} &=-\frac{1}{\mathrm{C}} \sum_{\mathrm{i}=1}^{\mathrm{C}} (\log \frac{\exp \left(\mathrm{P}_{\mathrm{c}_{\mathrm{i}}}^{\mathrm{S}} \cdot \tilde{P}_{c_{\mathrm{i}}}\right)}{\sum_{\mathrm{j}=1}^{\mathrm{C}} \exp \left(\mathrm{P}_{\mathrm{c}_{\mathrm{j}}}^{\mathrm{S}} \cdot \tilde{P}_{\mathrm{c}_{\mathrm{i}}}\right)} \\
&+ \log \frac{\exp \left(\mathrm{P}_{\mathrm{c}_{\mathrm{i}}}^{\mathrm{T}} \cdot \tilde{P}_{\mathrm{c}_{\mathrm{i}}}\right)}{\sum_{\mathrm{j}=1}^{\mathrm{C}} \exp \left(\mathrm{P}_{\mathrm{c}_{\mathrm{j}}}^{\mathrm{T}} \cdot \tilde{P}_{\mathrm{c}_{\mathrm{i}}}\right)}),
\end{aligned}
\end{equation}
where the dot product "\begin{math} \cdot \end{math}"  is used to measure the similarity between paired prototypes. \begin{math} C \end{math} denotes the total number of categories in the dataset. It is noteworthy that, similar to the computation of adversarial learning loss, we also need to mask the loss calculation for categories that are not present in the batch of training images. 

\subsection{Network Training} 

The DATR is trained with three loss functions: the supervised detection loss \begin{math} L_{det} \end{math}, the adversarial learning loss \begin{math} L_{adv} \end{math} as defined in Eq. (3), and the contrastive learning loss \begin{math} L_{contrast} \end{math} as defined in Eq. (6). In the Burn-In stage, the supervised detection loss is exclusively trained using data from the source domain. The training objective can be defined as follows:
\begin{equation}
\begin{aligned}
L &= L_{det}^{\text {sup }}  + \lambda_{a}L_{a d v} + \lambda_{c}L_{contrast}, \\
  &= L_{\text {det }}\left(P_{\text {src }}, Y_{\text {src }}\right)  + \lambda_{a}L_{a} + \lambda_{c}L_{contrast}, \\
\end{aligned}
\end{equation}
where \begin{math}P_{\text {src }}\end{math} represents the predicted bounding box for source data, \begin{math}Y_{\text {src }}\end{math} denotes the ground truth, and \begin{math} L_{det}^{\text {sup }} \end{math} denotes the supervised object detection loss, which remains consistent with DINO. The hyperparameters \begin{math} \lambda_{a} \end{math} and \begin{math} \lambda_{c} \end{math} are used to balance the supervised detection loss and other losses.

In the Teacher-Student Mutual Learning stage, images from the target domain are incorporated into supervised training by utilizing the generated pseudo-labels. Therefore, the training objective of DATR is defined as follows:
\begin{equation}
\begin{aligned}
L &= L_{det}^{\text {sup }} + \lambda_{unsup}L_{det}^{\text {unsup}} + \lambda_{a}L_{a d v} + \lambda_{c}L_{contrast}, \\
  &= L_{\text {det }}\left(P_{\text {src }}, Y_{\text {src }}\right) + \lambda_{unsup}L_{\text {det }}\left(P_{\text {tgt }}, Y_{\text {tgt }}\right) \\
  &+ \lambda_{a}L_{a d v} + \lambda_{c}L_{contrast}, \\
\end{aligned}
\end{equation}
where \begin{math}P_{\text {tgt }}\end{math} denotes the predicted bounding box for the target data, \begin{math}Y_{\text {src }}\end{math} is the generated pseudo labels, \begin{math} L_{det}^{\text {unsup }} \end{math} represents the unsupervised object detection loss, and \begin{math} \lambda_{unsup} \end{math} denotes the balancing weights for the corresponding
learning loss function.

\section{EXPERIMENTS}
This section details our experimentation, which includes datasets and evaluation metric, implementation details, comparisons with state-of-the-art approaches, as well as ablation studies and analysis. Detailed discussions on each of these aspects are provided in the subsequent subsections.

\subsection{Datasets and evaluation metric}

Following \cite{Wang_Cao_Zhang_He_Zha_Wen_Tao_2021,Yu_Liu_Wei_Zhou_Nakata_Gudovskiy_Okuno_Li_Keutzer_Zhang_2022,he2023bidirectional,zhao2023masked}, our proposed DATR is evaluated under three widely adopted domain adaptation scenarios, utilizing four datasets: Cityscapes, Foggy Cityscapes, Sim10k and BDD100k.

\textbf{Cityscapes}\cite{Cordts_Omran_Ramos_Rehfeld_Enzweiler_Benenson_Franke_Roth_Schiele_2016} is an urban scenes dataset and extensively used for evaluating cross-domain object detection performance. It encompasses 2,975 training images and 500 validation images, covering 50 cities across various seasons and times of the day. Consistent with other methods, our experiments focus on 8 categories within the dataset.

\textbf{Foggy Cityscapes}\cite{Sakaridis_Dai_Van_Gool_2018} is created by integrating fog into the original images from the Cityscapes. This process involved generating three fog densities (0.02, 0.01, 0.005), each corresponding to a specific range of visibility. Combined with Cityscapes, this dataset facilitates the evaluation of the method's effectiveness in knowledge transfer under adverse weather conditions. In our experiments, we focused on the identical eight categories as in Cityscapes, conducting evaluations on the images encompassing 0.02 fog densities.

\textbf{Sim10k}\cite{Johnson_Roberson_Barto_Mehta_Sridhar_Rosaen_Vasudevan_2016} is a synthetic image dataset generated by the Grand Theft Auto game engine, comprising 10,000 training images with 58,701 annotations of car bounding boxes. In our experiments, we utilized this dataset in conjunction with Cityscapes to evaluate synthetic to real adaptation. Owing to the dataset's exclusive focus on the car category, our use of the Cityscapes dataset was correspondingly narrowed to the car class, omitting all other categories.

\textbf{BDD100k}\cite{yu2020bdd100k} is a comprehensive driving dataset with diverse scenarios. Following existing methods, we utilize the daytime subset of BBD100k as the target domain data, comprising 36,278 training images and 5,258 validation images. In our experiments, we train on the annotated Cityscapes training set and the unlabeled daytime subset of BBD100k training set, and evaluate on the validation set.

In this paper, we evaluate the performance of our proposed DATR across three elaborate domain adaptation scenarios. Following \cite{Chen_Li_Sakaridis_Dai_Van_Gool_2018}, we employ Mean Average Precision (mAP) with a threshold of 0.5 as our evaluation metric.

\subsection{Implementation details}

We adopt DINO as the base detector. Our method compares with both CNN-based and transformer-based domain adaptive detection methods. To ensure a fair comparison, all methods aim to use ResNet-50 (pretrained on ImageNet\cite{Li_Fei-Fei_2009}) as the backbone network whenever possible. In all experiments, our model is trained for a maximum of 46 epochs, with the first 36 epochs designated as the Burn-In stage, followed by 10 epochs dedicated to the Teacher-Student Mutual Learning stage. In the Burn-In stage, following the implementation of DINO, we train our models using the Adam optimizer\cite{Kingma_Ba_2014} with a base learning rate of \begin{math} 2\times 10^{-4}    \end{math}, \begin{math} \beta_{1} = 0.9 \end{math} and the learning rate is decayed at the 30-th epoch by a factor of \begin{math} 0.1 \end{math}. In the Teacher-Student Mutual Learning stage, we train our models using the Adam optimizer with a base learning rate of \begin{math} 2\times 10^{-4} \end{math}. For all scenarios, we set the weight factors \begin{math} \lambda_{a} \end{math} and \begin{math} \lambda_{c} \end{math} in Eq. (7) and 8 to 0.1, and \begin{math} \lambda_{unsup} \end{math} in Eq. (8) to 1.0. The smoothing hyper-parameter in the Exponential Moving Average (EMA) is set to 0.999. We conduct each experiment on a NVIDIA A6000 GPUs with 48 GB of memory.

\subsection{Comparing with state-of-the-arts approaches}

\begin{table*}[htbp]
  \centering
  \caption{Experimental results (\%) of the weather adaptation scenario: Cityscapes → Foggy cityscapes.}
    \begin{tabular}{c|c|c|cccccccc|c}
    \toprule
    \multicolumn{12}{c}{\textbf{Cityscapes → Foggy Cityscapes}} \\
    \midrule
    Method & Type  & Backbone & person & rider & car    & truck & bus   & train & mcycle & bicycle & $mAP_{50}$ \\
    \midrule
    Source-DINO(ICLR'23)\cite{zhang2022dino} & \multirow{2}[0]{*}{Transformer} & \multirow{2}[0]{*}{resnet-50} &43.7              &44.6       &52.6       &22.1       &33.0       &21.1       &25.0       &42.0       &35.6  \\
    Oracle-DINO(ICLR'23)\cite{zhang2022dino} &             &       & 65.7     & 63.7    &  80.4     &44.3       & 67.5      &44.4       & 46.1      & 57.4      &58.7  \\
    \midrule
    DAF(CVPR'18)\cite{Chen_Li_Sakaridis_Dai_Van_Gool_2018} & CNN      & resnet-50 & 48.2      &48.8      &61.5     &22.6       &43.1       &20.2       &30.3       &42.1       &39.6  \\
    SWF(CVPR'19)\cite{Saito_Ushiku_Harada_Saenko_2019} & CNN      & resnet-50 &49.0       &49.0              &61.4       &23.9       &43.1       &22.9       &31.0       &45.2       &40.7  \\ GPA(CVPR'20)\cite{Xu_Wang_Ni_Tian_Zhang_2020}   & CNN      & resnet-50 &49.5       &46.7       &58.6              &26.4       &42.2       &32.3       &29.1       &41.8       &40.8  \\
    CRDA(CVPR'20)\cite{Xu_Zhao_Jin_Wei_2020}  & CNN      & resnet-50 &49.8       &48.4       &61.9             &22.3       &40.7       &30.0       &29.9       &45.4       &41.1  \\
    SFA(ACM MM'21)\cite{Wang_Cao_Zhang_He_Zha_Wen_Tao_2021} & Transformer      & resnet-50 &  46.5            &  48.6     &62.6       &25.1       & 46.2      &29.4       & 28.3      & 44.0      & 41.3 \\
    MTTrans(ECCV'22)\cite{Yu_Liu_Wei_Zhou_Nakata_Gudovskiy_Okuno_Li_Keutzer_Zhang_2022} & Transformer      & resnet-50 &47.7       &49.9              & 65.2      &25.8       & 45.9      & 33.8      & 32.6      & 46.5      & 43.4 \\
    AQT(IJCAI'22)\cite{Huang_Lu_Lin_Xie_Lin} & Transformer      & resnet-50 & 49.3              & 52.3      & 64.4      & 27.7      & 53.7      &  46.5     & 36.0      &  46.4     & 47.1 \\
    DA-DETR(CVPR'23)\cite{zhang2023detr} & Transformer      & resnet-50 & 49.9      &50.0      & 63.1      & 24.0      & 45.8      &37.5       & 31.6      & 46.3      & 43.5 \\
    BiADT(ICCV'23)\cite{he2023bidirectional} & Transformer      & resnet-50 & 50.7      &56.3      & 67.1      & 28.8      & 53.7      &\textbf{49.5}       & 38.8      & 50.1      & 49.4 \\
    CMT(CVPR'23)\cite{cao2023contrastive} & CNN      & VGG-16 & 45.9      &55.7      & 63.7     & \textbf{39.6}      & \textbf{66.0}      &38.8       & 41.4      & 51.2      & 50.3 \\
    MRT(ICCV'23)\cite{zhao2023masked} & Transformer      & resnet-50 & 52.8      &51.7      & 68.7     & 35.9      & 58.1     &54.5       & 41.0      & 47.1      & 51.2\\
    \midrule
    DATR & Transformer      & resnet-50 &\textbf{60.6}       &\textbf{59.2}              &\textbf{74.9}       &39.5      &62.1       &27.5       &\textbf{45.5}      &\textbf{53.5}       &\textbf{52.8}  \\
    \bottomrule
    \end{tabular}%
  \label{tab:1}%
\end{table*}%

\begin{table}[htbp]
  \centering
  \caption{Experimental results (\%) of the Synthetic-to-Real Adaptation scenario: SIM10k → Cityscapes.}
    \begin{tabular}{c|c|c|c}
    \toprule
    \multicolumn{4}{c}{\textbf{SIM10k → Cityscapes }} \\
    \midrule
    \multicolumn{1}{c|}{Method} &  Type     & Backbone & $mAP_{50}$ \\
    \midrule
    Source-DINO(ICLR'23)\cite{zhang2022dino} & \multirow{2}[0]{*}{Transformer}      & \multirow{2}[0]{*}{resnet-50} &52.6  \\
    Oracle-DINO(ICLR'23)\cite{zhang2022dino} &       &       &76.9  \\
    \midrule
    DAF(CVPR'18)\cite{Chen_Li_Sakaridis_Dai_Van_Gool_2018} &CNN       & resnet-50 & 49.8 \\
    SWF(CVPR'19)\cite{Saito_Ushiku_Harada_Saenko_2019} & CNN      & resnet-50 &50.5  \\
    GPA(CVPR'20)\cite{Xu_Wang_Ni_Tian_Zhang_2020}   & CNN      & resnet-50 &51.3  \\
    CRDA(CVPR'20)\cite{Xu_Zhao_Jin_Wei_2020}  &CNN       & resnet-50 &52.1  \\
    SFA(ACM MM'21)\cite{Wang_Cao_Zhang_He_Zha_Wen_Tao_2021} & Transformer      & resnet-50 &52.6  \\
    MTTrans(ECCV'22)\cite{Yu_Liu_Wei_Zhou_Nakata_Gudovskiy_Okuno_Li_Keutzer_Zhang_2022} &  Transformer     & resnet-50 &57.9  \\
    AQT(IJCAI'22)\cite{Huang_Lu_Lin_Xie_Lin} & Transformer      & resnet-50 &53.4  \\
    DA-DETR(CVPR'23)\cite{zhang2023detr} & Transformer      & resnet-50 &54.7  \\
    BiADT(ICCV'23)\cite{he2023bidirectional} & Transformer      & resnet-50 &55.8  \\
    MRT(ICCV'23)\cite{zhao2023masked} & Transformer      & resnet-50 &62.0  \\
    \midrule
    DATR & Transformer      & resnet-50 & \textbf{66.3}  \\
    \bottomrule
    \end{tabular}%
  \label{tab:2}%
\end{table}%

To demonstrate the effectiveness and generalization capability of the DATR, we evaluate the performance of our proposed across representative distinct domain adaptation scenarios: (1) Weather Adaptation, from Cityscapes to Foggy Cityscapes, involving training the models on the Cityscapes dataset and evaluating them on the Foggy Cityscapes dataset; (2) Synthetic-to-Real Adaptation, from Sim10k to Cityscapes, entailing training on the synthetic Sim10k dataset and testing on the real-world Cityscapes dataset; (3) Scene Adaptation, from Cityscapes to the daytime subset of BDD100k, where the models are trained on Cityscapes and tested on the daytime subset of BDD100k. In each scenario, we first present the performance of the base detector. “Source-DINO” represents the model trained on the source domain and evaluated on the target domain dataset. “Oracle-DINO” refers to the model trained and evaluated entirely within the target domain dataset. Then, we compare DATR with several state-of-the-art unsupervised domain adaptation methods, including both CNN-based and Transformer-based detectors.

\textbf{Weather Adaptation.} In this scenario, the visibility of objects in foggy images significantly decreases compared to normal conditions on the task Cityscapes → Foggy Cityscapes. As shown in Table \ref{tab:1}, the DATR achieves a mAP of \begin{math} 52.8\% \end{math}, significantly outperforming the baseline model and surpassing the state-of-the-art method by a margin of \begin{math} 1.6\% \end{math} in mAP. This demonstrates the effectiveness of our method in typical cross-domain scenarios.

\textbf{Synthetic-to-Real Adaptation.} Table \ref{tab:2} presents results from experiments on synthetic-to-real adaptation for the task Sim10k → Cityscapes. It is observed that DATR achieves the highest accuracy, with a mAP of \begin{math} 66.3\% \end{math}, and shows significant improvements over previous work. The experiments demonstrate DATR's powerful capability in addressing single-category cross-domain detection tasks.

\textbf{Scene Adaptation.} We evaluate the cross-scene adaptation for the task from Cityscapes to BDD100K-daytime. Table \ref{tab:3} presents the experimental results, where DATR achieves the highest mAP of \begin{math} 41.9\% \end{math}. The outcomes of this experiment compellingly demonstrate the generalization capabilities of our proposed.

\begin{table*}[htbp]
  \centering
  \caption{Experimental results (\%) of the Scene Adaptation scenario: Cityscapes → BDD100K-daytime.}
    \begin{tabular}{c|c|c|ccccccc|c}
    \toprule
    \multicolumn{11}{c}{\textbf{Cityscapes → BDD100K-daytime}} \\
    \midrule
    Method & Type  & Backbone & person & rider    & car   & truck & bus   & mcycle & bicycle & $mAP_{50}$ \\
    \midrule
    Source-DINO(ICLR'23) & \multirow{2}[0]{*}{Transformer} & \multirow{2}[0]{*}{resnet-50} &45.9             &31.6       &67.6       & 20.6      & 21.1      & 19.1      & 24.2      &32.8  \\
    Oracle-DINO(ICLR'23) &       &              & 70.0      &  52.2     & 84.9      &  64.5     & 64.3      &  46.8     &  52.5     & 62.2 \\
    \midrule
   DAF(CVPR'18)\cite{Chen_Li_Sakaridis_Dai_Van_Gool_2018} & CNN   & resnet-50 &  28.9     & 27.4      &  44.2          & 19.1      & 18.0      & 14.2      & 22.4      & 24.9 \\
    ICR-CCR-SW(CVPR'20)\cite{xu2020exploring} & CNN   & resnet-50 &  32.8           & 29.3      &  45.8     & 22.7      & 20.6      &  14.9     & 25.5      &27.4  \\
    EMP(ECCV'20)\cite{Hsu_Tsai_Lin_Yang_2020}   & CNN   & resnet-50 &  39.6     & 26.8             &  55.8     & 18.8      & 19.1      & 14.5      & 20.1      & 27.8 \\
   SFA(ACM MM'21)\cite{Wang_Cao_Zhang_He_Zha_Wen_Tao_2021} & Transformer & resnet-50 & 40.2      & 27.6             & 57.5      & 19.1      & 23.4      & 15.4      &  19.2     & 28.9 \\
MTTrans(ECCV'22)\cite{Yu_Liu_Wei_Zhou_Nakata_Gudovskiy_Okuno_Li_Keutzer_Zhang_2022} & Transformer & resnet-50 & 44.1             &  30.1     & 61.5      & 25.1      & 26.9      & 17.7      & 23.0      & 32.6 \\
    AQT(IJCAI'22)\cite{Huang_Lu_Lin_Xie_Lin} & Transformer & resnet-50 & 38.2      &  33.0     & 58.4      &  17.3     &  18.4            & 16.9      &  23.5     & 29.4 \\
    O2net(ACM MM'22)\cite{gong2022improving} & Transformer & resnet-50 & 40.4       &31.2       &58.6       &  20.4     & 25.0      & 14.9      &  22.7     & 30.5 \\
    BiADT(ICCV'23)\cite{he2023bidirectional} & Transformer & resnet-50 & 42.0       &34.5       &59.9       &  17.2     & 19.2      & 17.8      &  24.4     & 32.7 \\
    MRT(ICCV'23)\cite{zhao2023masked} & Transformer & resnet-50 &  48.4           & 30.9      &  63.7     & 24.7      &25.5       & 20.2      &  22.6     & 33.7 \\
    \midrule
    DATR  & Transformer      & resnet-50        &\textbf{57.7}       &\textbf{37.7}       & \textbf{75.8}      & \textbf{31.3}      & \textbf{35.3}      & \textbf{28.8}      &\textbf{26.8}       & \textbf{41.9} \\
    \bottomrule
    \end{tabular}%
  \label{tab:3}%
\end{table*}%

\subsection{Ablation studies }
\label{Ablation studies}

In this section, we first conducted ablation experiments by replacing or removing parts of the components to effectively analyze the contribution of each component in our proposed DATR. Next, we explore the performance of some variants of the proposed Class-wise Prototypes Alignment (CPA) module, primarily focusing on the methods of extracting class-wise prototypes, as introduced in Section \ref{Subsection B}. Finally, we demonstrate that DETR-based detectors can be effectively combined with a self-training framework for unsupervised cross-domain detection tasks. This combination can further mitigate the domain bias of the detector by leveraging the generated pseudo-labels. The experiments are conducted under the weather adaptation scenario and the experimental results on the validation data from Foggy Cityscapes are presented.

\textbf{Effectiveness of each component.} The results are shown in TABLE \ref{tab:Ablation}. Training DINO exclusively with data from the source domain presents significant challenges in achieving excellent results, attributable to domain shifts. “Backone-align” used as a fundamental implementation, refers to the alignment of output features from the CNN backbone, which leads to a  \begin{math} 6.9\% \end{math} improvement in mAP. Rows 3 and 4 demonstrate that using only the proposed CPA module or DAS effectively improves cross-domain detection results, achieving improvements in mAP of \begin{math} 8.1\% \end{math} and \begin{math} 6.2\% \end{math}, respectively. Furthermore, the results show that our method effectively complements "Backbone-align," primarily because our approach focuses on aligning instance-level features, which is independent of the image-level alignment performed by "Backbone-align". By employing all feature alignment methods "Backbone-align + CPA + DAS", the cross-domain detection performance increased from \begin{math} 35.6\% \end{math} to \begin{math} 48.7\% \end{math}, achieving an improvement of \begin{math} 12.3\% \end{math}. Ultimately, we implemented a self-training framework on DATR to further mitigate domain bias, resulting in a \begin{math} 4.1\% \end{math} improvement.

\begin{table}[htbp]
  \centering
  \caption{Ablation study of DATR}
    \begin{tabular}{rccccc}
    \toprule
    \multicolumn{1}{c}{Source-only} & Backbone-align & CPA & DAS & Self-training & $mAP_{50}$ \\
    \midrule
    \multicolumn{1}{c}{\begin{math} \surd  \end{math}}  &       &       &       &       & 35.6 \\
    \midrule
          & \begin{math} \surd  \end{math}     &       &       &       & 42.5 \\
          &       &\begin{math} \surd  \end{math}     &       &       & 43.7 \\
          &       &       & \begin{math} \surd  \end{math}     &       & 41.8 \\
          & \begin{math} \surd  \end{math}     & \begin{math} \surd  \end{math}    &       &       & 46.9 \\
          & \begin{math} \surd  \end{math}     &      & \begin{math} \surd  \end{math}    &       & 47.1 \\
          & \begin{math} \surd  \end{math}     & \begin{math} \surd  \end{math}     & \begin{math} \surd  \end{math}    &       & 48.7 \\
          & \begin{math} \surd  \end{math}     & \begin{math} \surd  \end{math}     & \begin{math} \surd  \end{math}     & \begin{math} \surd  \end{math}     & \textbf{52.8} \\
    \bottomrule
    \end{tabular}%
  \label{tab:Ablation}%
\end{table}%

\textbf{Extracting class-wise prototypes.} Ablation studies focusing on different methods for extracting class-wise prototypes are reported in Table \ref{tab:CPA}. The experiments were conducted on the basis of “Backbone-align”, and the variants of CPA module mainly include two different representative object queries selection criteria: based on confidence thresholds of detection results and the Hungarian matching algorithm, details are described in Section \ref{Subsection B}. Counterintuitively, filtered object queries did not improve cross-domain performance. We believe that the learning consistency of object queries may be disrupted due to involving only a subset of object queries in the alignment training process.

\begin{table}[htbp]
  \centering
  \caption{Ablation results of the different methods for extracting class-wise prototypes.}
    \begin{tabular}{rcccc}
    \toprule
    \multicolumn{1}{c}{Source-only} & Backbone-align & CPA   & Filtering method &$mAP_{50}$ \\
    \midrule
    \multicolumn{1}{c}{\begin{math} \surd  \end{math}} &       &       &—       & 35.6 \\
          & \begin{math} \surd  \end{math}     &       & —      &42.5  \\
    \midrule
          &     & \begin{math} \surd  \end{math}     &—       & 43.7 \\
           & \begin{math} \surd  \end{math}    & \begin{math} \surd  \end{math}     &—       &\textbf{46.9}  \\
          & \begin{math} \surd  \end{math}     & \begin{math} \surd  \end{math}     & Fixed threshold=0.2 & 41.4  \\
          & \begin{math} \surd  \end{math}     & \begin{math} \surd  \end{math}    & Fixed threshold=0.5 &44.8  \\
          & \begin{math} \surd  \end{math}     & \begin{math} \surd  \end{math}     & Fixed threshold=0.8 &44.0  \\
          & \begin{math} \surd  \end{math}     & \begin{math} \surd  \end{math}     & Hungarian matching &44.3  \\
    \bottomrule
    \end{tabular}%
  \label{tab:CPA}%
\end{table}%

\textbf{Exploring the impact of threshold values on self-training Framework.} We further analyze the hyperparameter related to the confidence threshold value, which is used to control the quality of pseudo-label generation within the self-training framework. Following the methodology established by Xu et al. \cite{Xu_Chen_Guan_Hu_Group}, we adjusted the range of threshold values from 0.2 to 0.7 to examine their impact on performance in the adaptation scenario from Cityscapes to the Foggy Cityscapes task. As shown in Table \ref{tab:threshold}, the self-training framework can effectively imporve the cross-domain detection performance of the DETR-Based detector, regardless of the threshold setting. Optimal threshold values indeed bring about further improvements. Based on our experimental results, we set the self-training threshold to 0.3 across all cross-domain scenarios.

\begin{table}[htbp]
  \centering
  \caption{Ablation studies of threshold value.}
    \begin{tabular}{ccccc}
    \toprule
    \multicolumn{2}{c}{Method} & \multicolumn{2}{c}{Threshold Value} & $mAP_{50}$ \\
    \midrule
    \multicolumn{2}{c}{DATR} & \multicolumn{2}{c}{—} & 48.7 \\
    \midrule
    \multicolumn{2}{c}{\multirow{6}[0]{*}{DATR with Self-training}} & \multicolumn{2}{c}{0.2} & 51.1 \\
    \multicolumn{2}{c}{} & \multicolumn{2}{c}{\textbf{0.3}} & \textbf{52.8} \\
    \multicolumn{2}{c}{} & \multicolumn{2}{c}{0.4} & 52.2 \\
    \multicolumn{2}{c}{} & \multicolumn{2}{c}{0.5} & 51.7 \\
    \multicolumn{2}{c}{} & \multicolumn{2}{c}{0.6} & 51.2 \\
    \multicolumn{2}{c}{} & \multicolumn{2}{c}{0.7} & 50.6 \\
    \bottomrule
    \end{tabular}%
  \label{tab:threshold}%
\end{table}%

\subsection{Visualization and Analysis}

\begin{figure}[htbp]
\centering
\includegraphics[width=8cm]{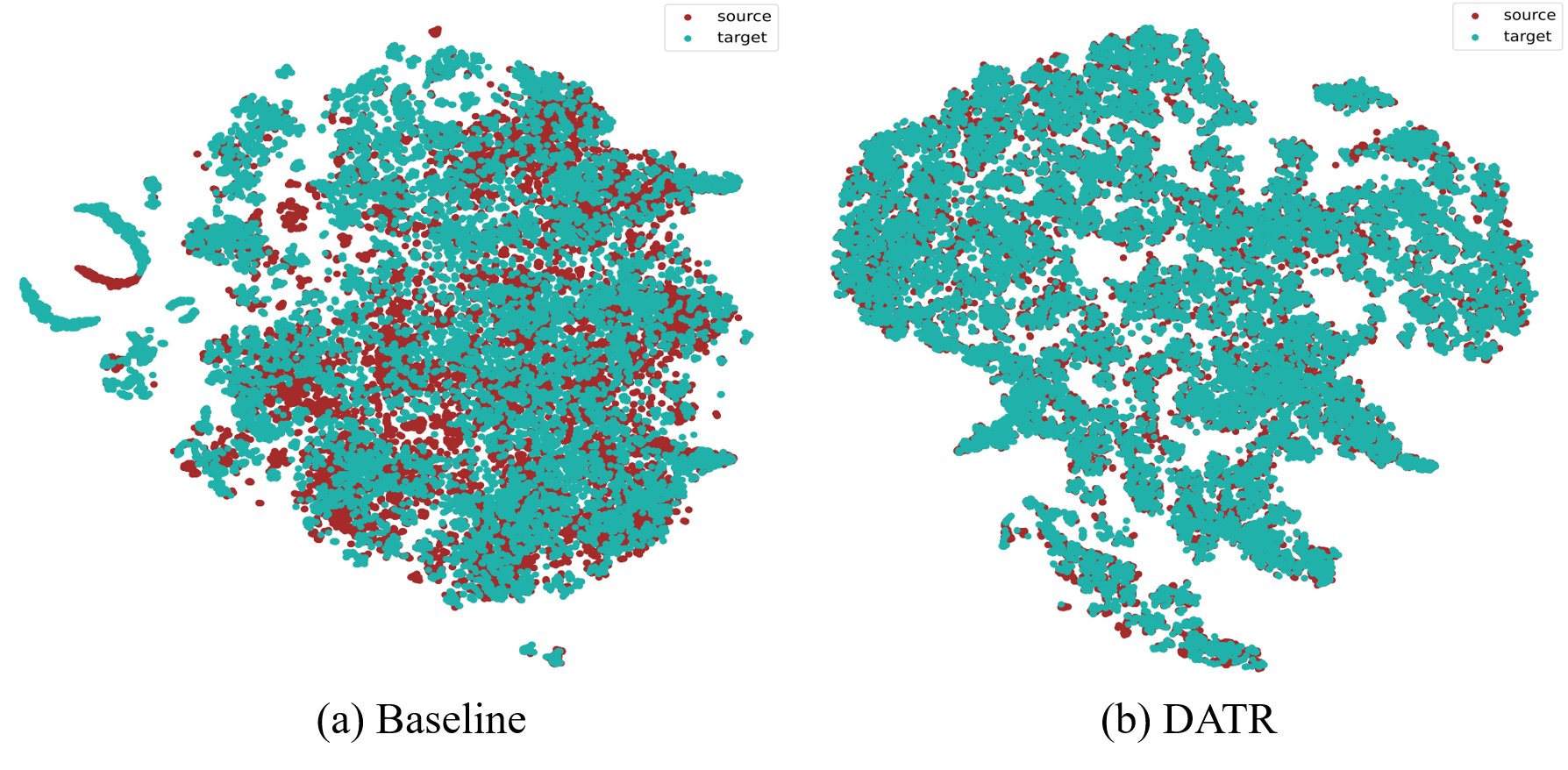}
\caption{The t-SNE visualization of object features from images originating from different domains. Our method aligns the domain shift well compared to the baseline method.}
\label{fig_4}
\end{figure}

\begin{figure}[htbp]
\centering
\includegraphics[width=8cm]{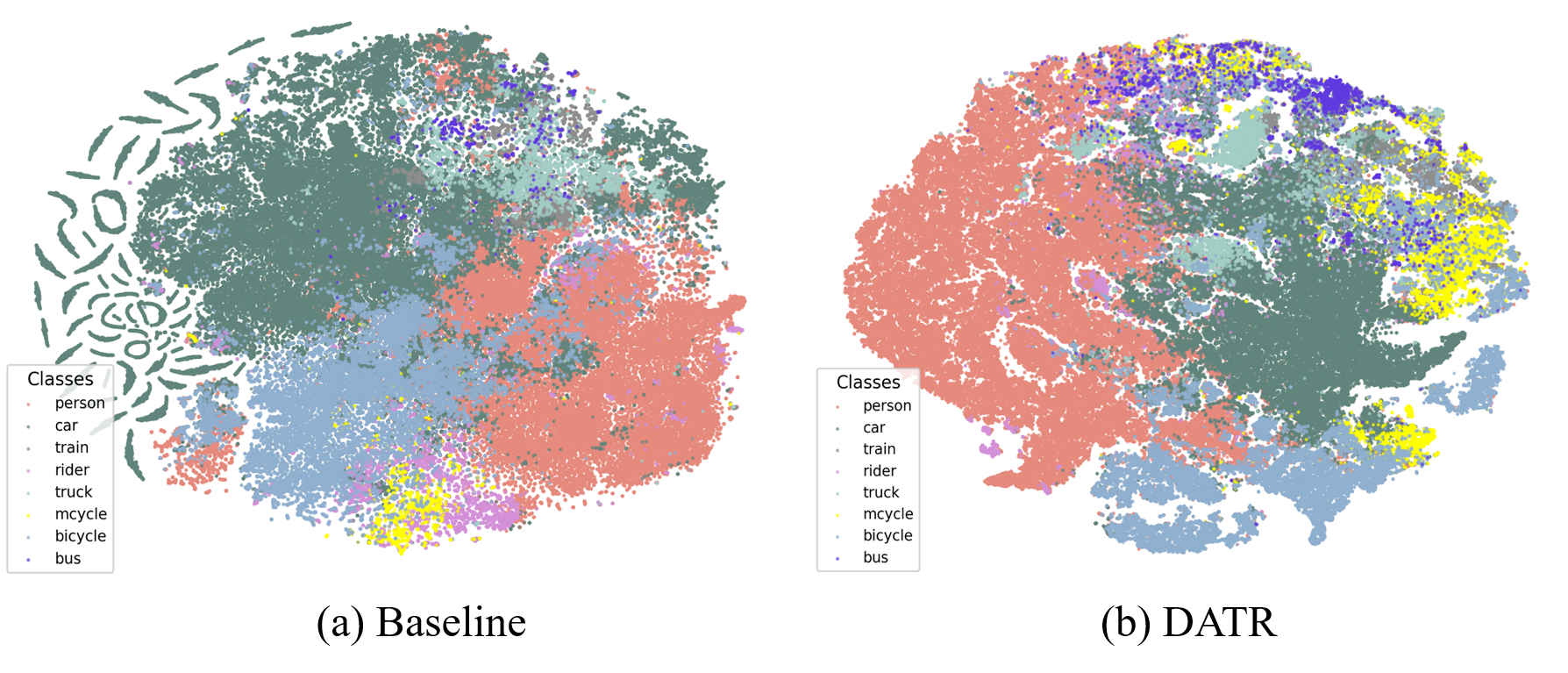}
\caption{The t-SNE visualization of object features that belong to eight object classes within the Foggy Cityscapes images. Our method enhances both the global representation and inter-class discriminability in the resultant feature space
}
\label{fig_5}
\end{figure}

\textbf{Feature visualization.} By utilizing the t-distributed stochastic neighbor embedding (t-SNE) method\cite{Maaten_Hinton_2008}, we visualized the features extracted from object queries in the task from Cityscapes to Foggy Cityscapes. Fig. \ref{fig_4} shows that our method exhibits a minimal domain gap compared to the baseline, which is trained solely on the source domain. This effect is primarily attributed to our proposed Class-wise Prototypes Alignment (CPA) module, which effectively aligns features from different domains in a class-aware manner. In Fig. \ref{fig_5}, we visualize object features by class category. Evidently, DATR enhances both the global representation and inter-class discriminability in the resultant feature space by utilizing our Dataset-level Alignment Scheme (DAS).

\begin{figure}[htbp]
\centering
\includegraphics[width=9cm]{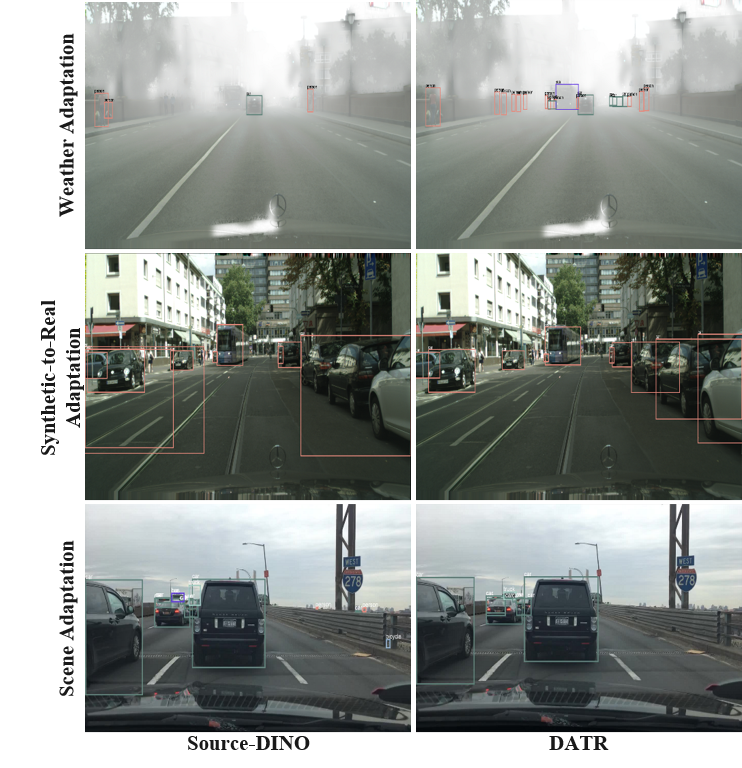}
\caption{Visualized results are provided across all experimental domain adaptation scenarios, with the confidence threshold for visualization set at 0.2. 'Source-DINO' represents the base detector that uses only source domain data for training.
}
\label{fig_6}
\end{figure}

\textbf{Detection results.} We present the visualization results of DATR across all experimental domain adaptation scenarios in Fig. \ref{fig_6}. Compared to baseline methods, our approach demonstrates more accurate detection results, including a reduction in false positives and the identification of challenging objects that the basic detector might overlook. The visual results correspond with the numerical evaluations, indicating that DATR exhibits exceptional performance and generalization capabilities across widely adopted domain adaptation scenarios. 

\section{Conclusion}

This paper introduces DATR, a powerful DETR-based detector designed for unsupervised domain adaptation in object detection. First, we present the Class Prototype Alignment (CPA) module, designed to effectively align features in a class-aware manner by establishing a linkage between detection tasks and domain adaptation tasks. Subsequently, we introduce a Dataset-Level Alignment Scheme (DAS) designed to optimize the detector's feature representation at the dataset level by utilizing contrastive learning, thereby enhancing the model's cross-domain detection performance. Furthermore, the DATR adopts a mean-teacher self-training framework to further mitigate the bias across different domains. Comprehensive experiments conducted across various domain adaptation scenarios have shown that DATR exhibits superior performance in unsupervised domain adaptation for object detection tasks. In future work, we plan to investigate methods that can enhance cross-domain object detection performance, even with a limited number of samples available.

%\begin{thebibliography}{1}
\bibliographystyle{IEEEtran}
\bibliography{References.bib}
%\end{thebibliography}

\vfill

\end{document}